%% file: acl_latex.tex
\documentclass[11pt]{article}
\ifdefined\pdfminorversion
\fi

\usepackage[preprint]{acl}

\usepackage{times}
\usepackage{latexsym}
\usepackage[T1]{fontenc}
\usepackage[utf8]{inputenc}
\usepackage{microtype}
\usepackage{inconsolata}
\usepackage{graphicx}

\usepackage{multirow}
\usepackage{amsmath}
\usepackage{booktabs}
\usepackage{amssymb}
\usepackage{makecell}
\usepackage{tabularx}
\usepackage{ragged2e}

\DeclareMathSizes{7.5}{8}{5}{5}

\newcommand{\NA}{\rule[0.5ex]{0.7em}{0.5pt}}

\usepackage{silence}
\title{X-GS: An Extensible Framework for Perceiving and Thinking with 3D Gaussian Splatting}

\author{
Yueen Ma\textsuperscript{\rm 1,\rm 3},
Zenglin Xu\textsuperscript{\rm 2,\rm 3},
Irwin King\textsuperscript{\rm 1}\\
\textsuperscript{\rm 1}The Chinese University of Hong Kong\\
\textsuperscript{\rm 2}Fudan University\\
\textsuperscript{\rm 3}Shanghai Academy of AI for Science\\
{\normalfont\normalsize
\texttt{mayueen@link.cuhk.edu.hk},\quad
\texttt{zenglinxu@fudan.edu.cn},\quad
\texttt{king@cse.cuhk.edu.hk}}
}

\begin{document}

\maketitle

\begin{abstract}
3D Gaussian Splatting (3DGS) has emerged as a powerful technique for novel view synthesis, subsequently extending into numerous spatial AI applications. However, most existing 3DGS methods operate in isolation, focusing on specific domains. In this paper, we introduce X-GS, an extensible framework that integrates previously isolated 3DGS methods into the perception module of a VLM for spatial tasks, with two major components: the \textit{Perceiver} and the \textit{Thinker}. The \textit{Perceiver} performs online 3DGS-based SLAM with semantic distillation and outputs semantic Gaussians from unposed video streams. It leverages recent vision foundation models for stronger geometric priors, and we introduce three novel optimizations to improve semantic distillation efficiency. The \textit{Thinker} interfaces diverse VLMs with these semantic Gaussians, unlocking spatial multimodal capabilities such as 3D visual grounding and scene captioning. Experimental results on diverse benchmarks demonstrate the efficiency and newly unlocked multimodal capabilities of the X-GS framework. Code is available at \url{https://github.com/yueen-ma/X-GS}.
\end{abstract}

\input{sections/1_intro}

\input{sections/2_related}

\input{sections/3_method}

\input{sections/4_experiment}

\section{Conclusion}
We propose X-GS, an extensible framework that integrates previously isolated 3DGS methods into the perception module of a VLM for spatial tasks through two components: the \textit{Perceiver} and the \textit{Thinker}. Our instantiation of the \textit{Perceiver} is an online 3DGS-based SLAM system with semantic distillation that reconstructs queryable semantic Gaussians from unposed video, made efficient by an online VQ module, GPU-accelerated grid-sampled supervision, and a parallelized pipeline. The \textit{Thinker} connects these Gaussians to contrastive VLMs for 3D visual grounding and to generative VLMs for scene captioning and question answering. Experiments show that X-GS outperforms monocular baselines in online reconstruction, matches or surpasses state-of-the-art semantic 3DGS methods in 3D visual grounding at a fraction of their computational cost, and improves on its base VLM in zero-shot 3D vision-language tasks; after finetuning, it outperforms specialized 3D VLMs.

\section*{Limitations}
First, X-GS currently consists of two main modules: the \textit{Perceiver} and the \textit{Thinker}. Future work could explore end-to-end architectures. Second, as demonstrated by recent systems such as FreeSplat \cite{DBLP:conf/nips/WangHCL24}, SplatTalk~\cite{DBLP:journals/corr/abs-2503-06271}, EA3D~\cite{DBLP:journals/corr/abs-2510-25146}, and EmbodiedSplat~\cite{lee2026embodiedsplat}, there is a growing trend toward feed-forward architectures for 3DGS. Integrating such feed-forward mechanisms into X-GS may achieve faster speeds than the current optimization-based pipeline. However, such models also entail significant training costs for different VFMs. Finally, accommodating dynamic scenes using techniques such as 4DGS \cite{DBLP:conf/cvpr/WuYFX0000W24, ma2026fourdgswam} represents a highly promising direction for embodied AI applications \cite{DBLP:journals/corr/abs-2307-15818} and world models \cite{DBLP:journals/corr/abs-2603-19312}.

\section*{Acknowledgements}
The research presented in this paper was partially supported by the Research Grants Council of the Hong Kong Special Administrative Region, China (CUHK 2300246, RGC C1043-24G), (CUHK 14203425, RGC GRF 2151317), and CUHK 7010870.

\bibliography{references}

\clearpage

\appendix
\input{sections/5_appendix}

\end{document}

%% file: sections/1_intro.tex
\section{Introduction}
\label{sec:intro}

\input{figures/0_intro_fig}

The groundbreaking success of 3D Gaussian Splatting (3DGS)~\cite{DBLP:journals/tog/KerblKLD23} in real-time novel view synthesis has sparked rapid advancements across several distinct research directions. To overcome the reliance on computationally expensive offline Structure-from-Motion (SfM) pipelines like COLMAP, recent methods~\cite{DBLP:conf/cvpr/Fu0LKKE24} can reconstruct scenes directly from unposed image sequences. Concurrently, the advent of 3DGS-based SLAM systems~\cite{DBLP:conf/cvpr/MatsukiMKD24} has improved upon traditional point-cloud-based methods, allowing the mapped scene to retain significantly richer visual fidelity. Extending beyond pure appearance, semantic-GS~\cite{DBLP:journals/corr/abs-2507-07136} methods endow the 3D Gaussians with semantic feature channels to facilitate complex scene understanding. Finally, to bridge the gap between spatial representations and multimodal reasoning, the latest works have introduced Vision-Language Models (VLMs)~\cite{DBLP:journals/corr/abs-2503-06271} that natively process 3D Gaussians as visual inputs.

Despite this rapid progress, these advancements have typically been developed in isolation. To combine their respective strengths, we introduce X-GS, an extensible framework designed to integrate these disparate 3DGS techniques into the perception module of a VLM for spatial tasks, with two major components: the \textit{Perceiver} and the \textit{Thinker}, as illustrated in Figure~\ref{fig:intro}. The name stands for an e\textbf{\underline{X}}tensible framework for 3D\textbf{\underline{GS}}, where the ``X'' serves as a placeholder prefix denoting distinct paradigms such as pose-free-GS, VFM-for-GS, SLAM-GS, semantic-GS, and VLM-for-GS. By synthesizing these diverse capabilities, X-GS achieves online 3DGS-based SLAM with semantic distillation while simultaneously bridging the gap to downstream multimodal VLMs. Importantly, X-GS serves as a foundational blueprint; while this paper presents one concrete instantiation, its extensible nature ensures that as individual research directions evolve, advanced variants of X-GS can be readily integrated.

To realize this integrated vision using currently available 3DGS methods, our concrete instantiation of the framework is structured around these two components. The first, \textbf{X-GS-\textit{Perceiver}}, is an online 3DGS-based SLAM system with semantic distillation. It jointly models the camera pose, geometry, appearance, and semantics of an environment from unposed video streams and outputs semantic Gaussians. Functioning as an online SLAM system, the \textit{Perceiver} leverages recent Vision Foundation Models (VFMs)~\cite{DBLP:conf/cvpr/WangCKV0N25} to extract robust camera poses and dense geometric priors, which are then used to seed the 3D Gaussians. Simultaneously, it distills high-dimensional semantic features directly into these Gaussians. To reduce the computational cost of this semantic distillation during online mapping, we introduce three efficiency optimizations: an EMA-updated vector quantization (VQ) module, GPU-accelerated grid sampling for semantic supervision, and a meticulously parallelized system pipeline.

The second component, \textbf{X-GS-\textit{Thinker}}, is a versatile module that interfaces diverse VLMs with the semantic Gaussians output by the \textit{Perceiver}, unlocking spatial multimodal capabilities. For contrastive VLMs, the \textit{Thinker} directly queries the distilled features to facilitate open-vocabulary tasks like 3D visual grounding. For generative VLMs, it leverages the global 3D context to guide the model toward the most informative visual tokens across long video sequences. This effectively reduces temporal redundancy, supporting complex multimodal reasoning tasks such as 3D scene captioning and question answering.

In summary, our main contributions are:
\begin{itemize}
    \item We present X-GS, an extensible framework that integrates previously isolated 3DGS methodologies into the perception module of a VLM for spatial tasks, with two major components: the \textit{Perceiver} and the \textit{Thinker}.
    \item We propose X-GS-\textit{Perceiver}, an online 3DGS-based SLAM system with semantic distillation that outputs semantic Gaussians from unposed video streams; its efficiency is improved by three optimizations: an online VQ module, grid-sampled supervision, and a parallelized pipeline.
    \item We introduce X-GS-\textit{Thinker}, a versatile module that interfaces diverse VLMs with the semantic Gaussians output by the \textit{Perceiver}, unlocking spatial multimodal capabilities.
    \item Extensive experiments demonstrate that our integrated framework achieves state-of-the-art performance in both online semantic 3D mapping and downstream 3D reasoning tasks.
\end{itemize}

%% file: figures/0_intro_fig.tex
\begin{figure}[t]
\centering
\includegraphics[width=1.0\columnwidth, trim={0.688in 10.74in 4.188in 0.5in},clip]{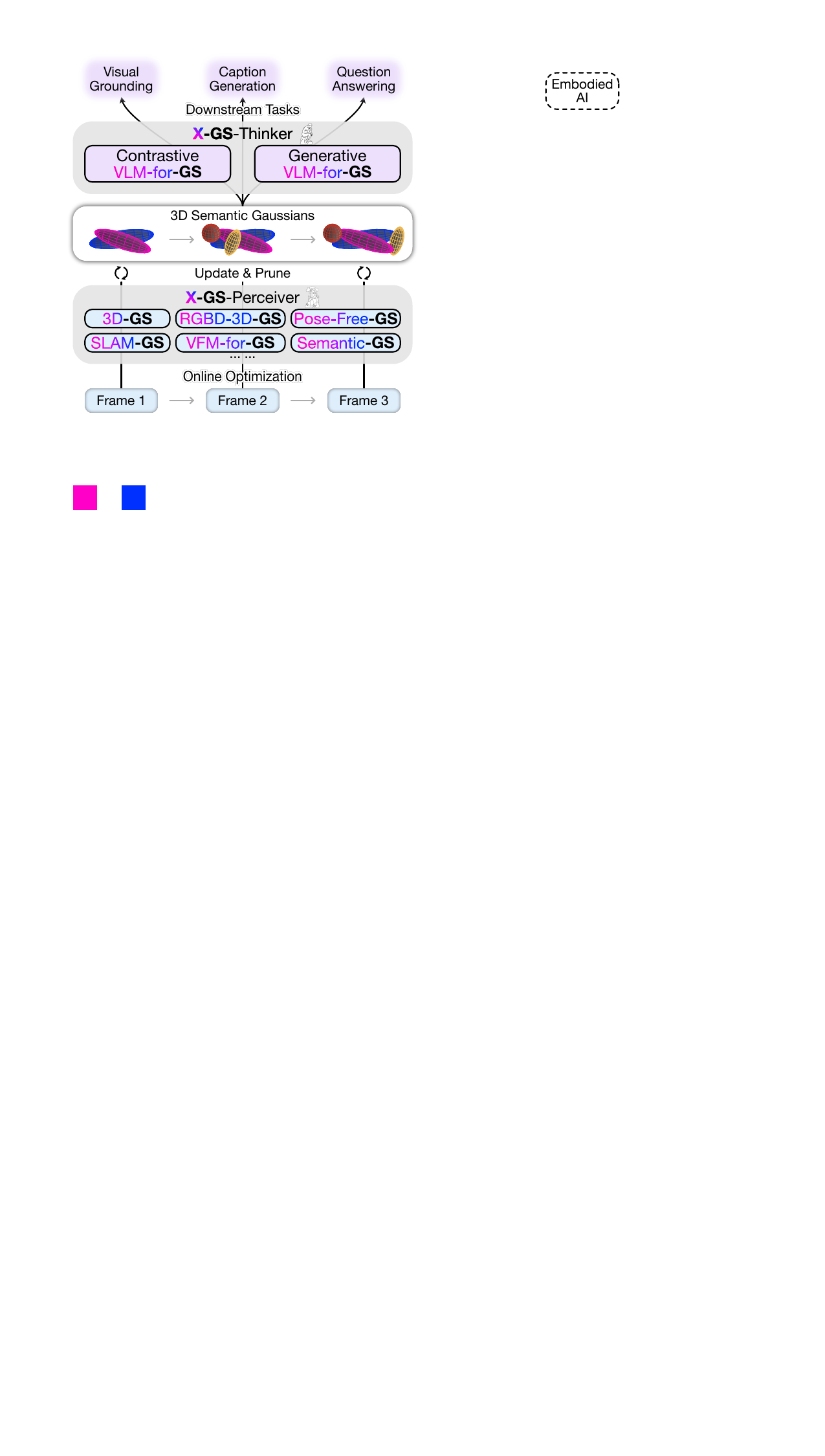}
\caption{X-GS is an extensible framework that integrates previously isolated 3DGS methodologies into the perception module of a VLM for spatial tasks, with two major components: the \textit{Perceiver} and the \textit{Thinker}. X-GS-\textit{Perceiver} is an online 3DGS-based SLAM system with semantic distillation that outputs semantic Gaussians. X-GS-\textit{Thinker} interfaces diverse VLMs with these semantic Gaussians, unlocking spatial multimodal capabilities.}
\label{fig:intro}
\end{figure}

%% file: sections/2_related.tex
\section{Related Work}

\input{figures/1_overview_fig}
\input{tables/0_comparison.tex}

\subsection{3DGS-based Online SLAM}
Standard 3D Gaussian Splatting (3DGS) algorithms~\cite{DBLP:journals/tog/KerblKLD23} rely on offline multi-view image sequences and COLMAP initialization. While subsequent methods like CF-3DGS~\cite{DBLP:conf/cvpr/Fu0LKKE24} eliminate the need for pre-computed poses, they still require offline processing. Recently, approaches such as MonoGS~\cite{DBLP:conf/cvpr/MatsukiMKD24}, GS-SLAM~\cite{DBLP:conf/cvpr/YanQXZWW024}, and Gaussian-SLAM~\cite{DBLP:journals/corr/abs-2312-10070} have achieved online SLAM using 3DGS. However, these systems focus exclusively on 3D reconstruction and camera tracking, ignoring high-level semantic scene understanding. To enrich 3DGS representations, Feature 3DGS~\cite{DBLP:conf/cvpr/ZhouCJFZXCYWK24} and LangSplat~\cite{DBLP:conf/cvpr/Qin0ZWP24, DBLP:journals/corr/abs-2507-07136} distill dense semantic features from vision foundation models (e.g., SAM or CLIP) directly into the 3D Gaussian field, enabling language-driven queries and zero-shot 3D object detection. Nevertheless, these semantic methods rely heavily on precise, pre-computed camera poses and are strictly designed for offline mapping, inherently limiting their applicability in dynamic or autonomous environments.

\subsection{Semantic SLAM}
Existing semantic 3DGS SLAM systems, such as LEGO-SLAM~\cite{DBLP:journals/corr/abs-2511-16144} and OpenMonoGS-SLAM~\cite{DBLP:journals/corr/abs-2512-08625}, attempt online semantic mapping but suffer from rigid architectural bottlenecks. LEGO-SLAM strictly requires RGB-D input, whereas OpenMonoGS-SLAM operates solely on monocular video—lacking the flexibility to utilize depth maps even when available. Furthermore, OpenMonoGS-SLAM is computationally expensive. X-GS natively overcomes all of these limitations.

\subsection{VLMs for 3DGS}
Recent literature~\cite{world-labs-about} demonstrates the immense value of explicitly integrating 3D representations with Vision-Language Models (VLMs) for downstream reasoning. For instance, UniGS~\cite{DBLP:conf/iclr/LiZTSZKXL25} aligns optimized 3D Gaussians with textual spaces for multimodal contrastive learning~\cite{DBLP:conf/aaai/MaSHLZK23}, while SplatTalk~\cite{DBLP:journals/corr/abs-2503-06271} and ChatSplat~\cite{DBLP:journals/corr/abs-2412-00734} build VLMs that take 3DGS as visual input. However, current VLMs for 3DGS are restricted entirely to static, offline scenes. X-GS integrates these disjoint paradigms into a single framework.

Table~\ref{tab:related_work_comparison} provides a comparison of key capabilities among representative 3DGS methods. To the best of our knowledge, X-GS is the first framework to fulfill all the listed requirements.

%% file: figures/1_overview_fig.tex
\begin{figure*}[t]
\centering
\includegraphics[width=1.0\textwidth, trim={0.437in 10.875in 0.437in 1.0in},clip]{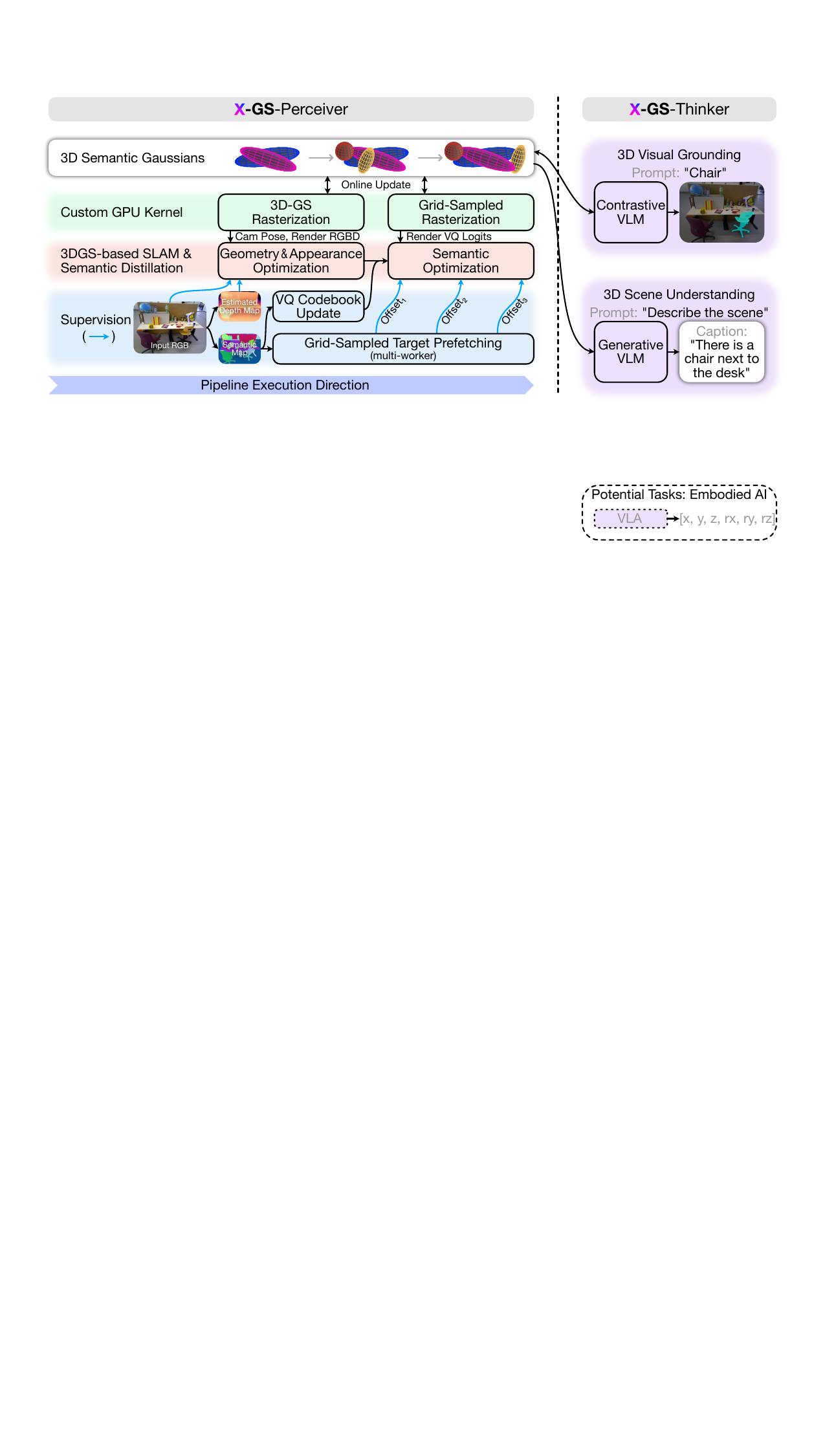}
\caption{\textbf{Overview of the X-GS framework.} The X-GS-\textit{Perceiver} is an online 3DGS-based SLAM system with semantic distillation that outputs semantic Gaussians; a Vector Quantization (VQ) module, grid-based semantic supervision, and an asynchronous parallel pipeline improve its efficiency. Designed as a versatile framework, it can flexibly incorporate various Vision Foundation Models (VFMs). Furthermore, the X-GS-\textit{Thinker} interfaces diverse VLMs with these semantic Gaussians, unlocking spatial multimodal capabilities such as 3D grounding and reasoning.}
\label{fig:overview}
\end{figure*}

%% file: tables/0_comparison.tex
\begin{table}[t]
\centering
\caption{\textbf{Comparison of X-GS with representative 3DGS methods.}}
\label{tab:related_work_comparison}

\newcommand{\cmark}{\textcolor{green!65!black}{$\checkmark$}}
\newcommand{\xmark}{\textcolor{red}{$\boldsymbol{\times}$}}
\newcommand{\colhead}[1]{\rotatebox{90}{\textbf{#1}}}

\setlength{\tabcolsep}{4.5pt}
\small
{\microtypesetup{protrusion=false}%
\resizebox{\columnwidth}{!}{%
\begin{tabular}{l *{5}{c} c@{\hspace{10pt}}}
\toprule
\textbf{Method} &
    \colhead{RGB-Only} &
    \colhead{RGB-D} &
    \colhead{Pose-Free} &
    \colhead{Online-SLAM} &
    \colhead{Semantics} &
    \colhead{Generative VLM} \\
\midrule
\textit{\textbf{Offline 3DGS}} \\
3DGS~\cite{DBLP:journals/tog/KerblKLD23}         & \cmark & \xmark & \xmark & \xmark & \xmark & \xmark \\
CF-3DGS~\cite{DBLP:conf/cvpr/Fu0LKKE24}      & \cmark & \xmark & \cmark & \xmark & \xmark & \xmark \\
Feature 3DGS~\cite{DBLP:conf/cvpr/ZhouCJFZXCYWK24} & \cmark & \xmark & \xmark & \xmark & \cmark & \xmark \\
LangSplat~\cite{DBLP:conf/cvpr/Qin0ZWP24}    & \cmark & \xmark & \xmark & \xmark & \cmark & \xmark \\
LangSplatV2~\cite{DBLP:journals/corr/abs-2507-07136}  & \cmark & \xmark & \xmark & \xmark & \cmark & \xmark \\
\midrule
\textit{\textbf{Online 3DGS}} \\
MonoGS~\cite{DBLP:conf/cvpr/MatsukiMKD24}      & \cmark & \cmark & \cmark & \cmark & \xmark & \xmark \\
LEGO-SLAM~\cite{DBLP:journals/corr/abs-2511-16144}    & \xmark & \cmark & \cmark & \cmark & \cmark & \xmark \\
OpenMonoGS-SLAM~\cite{DBLP:journals/corr/abs-2512-08625} & \cmark & \xmark & \cmark & \cmark & \cmark & \xmark \\
EA3D~\cite{DBLP:journals/corr/abs-2510-25146}      & \cmark & \xmark & \cmark & \cmark & \cmark & \xmark \\
EmbodiedSplat~\cite{lee2026embodiedsplat}          & \cmark & \cmark & \xmark & \cmark & \cmark & \xmark \\
\midrule
\textit{\textbf{VLMs for 3DGS}} \\
UniGS~\cite{DBLP:conf/iclr/LiZTSZKXL25}        & \cmark & \xmark & \xmark & \xmark & \cmark & \xmark \\
SplatTalk~\cite{DBLP:journals/corr/abs-2503-06271}    & \cmark & \xmark & \xmark & \xmark & \cmark & \cmark \\
\midrule
\textbf{X-GS (Ours)}       & \textbf{\cmark} & \textbf{\cmark} & \textbf{\cmark} & \textbf{\cmark} & \textbf{\cmark} & \textbf{\cmark} \\
\bottomrule
\end{tabular}%
}%
}
\end{table}

%% file: sections/3_method.tex
\section{Method}

We propose X-GS, an extensible framework designed to integrate previously isolated 3DGS advancements into the perception module of a VLM for spatial tasks, with two major components: the \textit{Perceiver} and the \textit{Thinker}. As illustrated in Figure~\ref{fig:overview}, \textbf{X-GS-\textit{Perceiver}} is an online 3DGS-based SLAM system with semantic distillation: it ingests unposed monocular video streams (with optional depth) to continuously co-optimize 3D Gaussians and camera poses while distilling VFM semantic features into them, outputting semantic Gaussians. \textbf{X-GS-\textit{Thinker}} is a versatile module that interfaces diverse VLMs with these semantic Gaussians, unlocking spatial multimodal capabilities such as text-prompted 3D visual grounding and zero-shot scene captioning.

\subsection{Preliminaries}

3DGS represents a scene using 3D Gaussians, each parameterized by $\Theta = \{\boldsymbol{\mu}, \boldsymbol{\Sigma}, \alpha, \mathbf{c}\}$ (center, covariance, opacity, and color). For a camera pose $\mathbf{T} \in \text{SE}(3)$, each 3D Gaussian is projected into a 2D elliptical splat; the expected pixel color $C(x)$ is then computed by front-to-back $\alpha$-blending $N_g$ depth-sorted splats:
\begin{equation}\label{eq:rendering}
C(x \mid \mathbf{T}, \Theta) = \sum_{i=1}^{N_g} \mathbf{c}_i \alpha'_i \prod_{j=1}^{i-1} (1 - \alpha'_j),
\end{equation}
where $\alpha'_i$ is the evaluated 2D opacity of the $i$-th splat. This rasterization naturally extends to semantic features: by augmenting the 3D Gaussians with a set of $D$-dimensional features $\mathbf{f} = \{\mathbf{f}_i\}_{i=1}^{N_g}$, we render a dense 2D semantic feature map $F(x \mid \mathbf{T}, \Theta, \mathbf{f})$ by substituting $\mathbf{c}_i$ with $\mathbf{f}_i$ in the identical blending formulation.

\subsection{X-GS-\textit{Perceiver}}
X-GS-\textit{Perceiver} is designed to jointly model the geometry, appearance, and semantics of an environment, yielding a comprehensive 3D representation that facilitates various downstream tasks. It is highly extensible, as its semantic features can be distilled from a wide range of Vision Foundation Models (VFMs), such as SAM, CLIP, and SigLIP.

\input{figures/2_deps_fig}

\input{tables/1_slam}
\input{figures/3_qualitative_fig}

\subsubsection{3DGS-based SLAM}
3DGS was originally designed to process multi-view images offline. MonoGS \cite{DBLP:conf/cvpr/MatsukiMKD24} later proposed converting 3DGS into an online SLAM pipeline, operating via two main phases: tracking and mapping. During tracking, camera poses are estimated. During mapping, the 3D Gaussians are optimized to match the input images.

Unlike MonoGS, we leverage recent VFMs that can simultaneously provide accurate spatial information, including camera poses, depth estimates, and point clouds, exemplified by VGGT \cite{DBLP:conf/cvpr/WangCKV0N25}. Since the original VGGT is also an offline model, we employ its online variants, such as StreamVGGT \cite{DBLP:journals/corr/abs-2507-11539} and InfiniteVGGT \cite{DBLP:journals/corr/abs-2601-02281}.

\paragraph{Tracking.}
During tracking, for each incoming frame, we use the online VFM's camera pose estimation for initialization, replacing the zero-velocity assumption used in MonoGS. Although these VFMs provide robust initial camera poses, the poses may still accumulate slight drift over time. Therefore, we further finetune the poses against the input color image $I_t$ and depth map $D_t$ by minimizing the tracking loss $\mathcal{L}_{\text{track}}$:
\begin{equation}\label{eq:tracking}
\mathbf{T}_t^* = \underset{\mathbf{T}_t}{\arg\min} \ \mathcal{L}_{\text{track}}(I_t, D_t \mid \mathbf{T}_t, \Theta).
\end{equation}
Depth supervision is primarily supplied by the VFM, though we can seamlessly accommodate ground-truth input depth maps if available.

\paragraph{Mapping.}
The mapping phase refines the global scene representation by co-optimizing the Gaussian parameters $\Theta$ along with the camera poses $\mathbf{T}$. Specifically, we first initialize new Gaussians from point clouds provided directly by the VFMs or backprojected from their estimated depth maps. This serves as a much stronger geometric prior than naive flat-plane initialization. Furthermore, to mitigate rendering artifacts, we incorporate an occluder pruning mechanism that removes floating Gaussians that are much closer to the camera than the estimated depth, as they can occlude the Gaussians behind and hinder the optimization process.

The mapping optimization is performed over a sliding window of keyframes $\mathcal{K}$. Instead of relying on heuristic 2D overlap ratios, the window is dynamically maintained by inserting new keyframes based on the VFM's robust 4D point tracks and visibility masks, while evicting redundant older views to preserve a maximum capacity of $N_{win}$. The mapping thread refines $\Theta$ by minimizing the mapping loss $\mathcal{L}_{\text{map}}$ across this active window of ground-truth images $I_k$ and depth maps $D_k$:
\begin{equation}\label{eq:mapping}
\Theta^* = \underset{\Theta}{\arg\min} \sum_{k \in \mathcal{K}} \mathcal{L}_{\text{map}}(I_k, D_k \mid \Theta, \mathbf{T}_k).
\end{equation}
To better preserve the VFM's depth map prior, we slow the learning rate for the Gaussians' position updates during this optimization.

\subsubsection{Online Semantic Distillation}

Geometry and appearance information provided by plain 3DGS cannot satisfy the needs of downstream multimodal tasks. Therefore, we add semantic distillation capabilities to the SLAM pipeline. Naively attaching semantic features to 3D Gaussians has proven to be heavily memory- and compute-intensive \cite{DBLP:conf/cvpr/Qin0ZWP24}. To reduce this cost during online mapping, we introduce three efficiency optimizations: an online Vector Quantization (VQ) module, a GPU-accelerated grid-sampling scheme, and a meticulously designed pipeline that fully leverages highly parallelized scheduling. Figure~\ref{fig:deps} illustrates the information flow among these components and the grid-sampled semantic supervision scheme.

\paragraph{Online VQ Module.}
Vector Quantization reduces the dimensionality of semantic features at each pixel. Inspired by LangSplatV2 \cite{DBLP:journals/corr/abs-2507-07136}, we represent the semantic state of each Gaussian $i$ using a learnable logit vector $\mathbf{z}_i \in \mathbb{R}^{L}$ over a shared codebook $\mathbf{E} \in \mathbb{R}^{L \times D}$, where the $l$-th row $\mathbf{E}_{l, *} \in \mathbb{R}^{D}$ represents a single codeword. These logits are converted into mixture weights via $\mathbf{w}_i = \operatorname{softmax}(\mathbf{z}_i)$. The decoded semantic feature is then obtained by a weighted sum over the codebook: $\hat{\mathbf{f}}_i = \mathbf{E}^\top \mathbf{w}_i \in \mathbb{R}^{D}$. This parameterization separates Gaussian-specific coefficients from shared semantic codewords.

We update the shared codebook online using an exponential moving average (EMA). Given a set of observed semantic features $\{\mathbf{x}_n\}_{n=1}^{N}$, each is assigned to its nearest codeword $a_n = \arg\min_l \| \mathbf{x}_n - \mathbf{E}_{l, *} \|_2$. We accumulate the assignment counts $c_l = \sum_n \mathbb{I}[a_n = l]$ and feature sums $\mathbf{s}_l = \sum_n \mathbb{I}[a_n = l] \mathbf{x}_n$. The codebook is then updated via EMA buffers, where $N_l$ tracks the moving cluster size and the matrix $\mathbf{M} \in \mathbb{R}^{L \times D}$ tracks the moving feature sums. For each codeword $l$, the corresponding EMA statistics and codebook row are updated:
\begin{equation}\label{eq:ema_update}
\begin{aligned}
N_l &\leftarrow \lambda N_l + (1-\lambda)c_l, \\
\mathbf{M}_{l, *} &\leftarrow \lambda \mathbf{M}_{l, *} + (1-\lambda)\mathbf{s}_l, \\
\mathbf{E}_{l, *} &\leftarrow \frac{\mathbf{M}_{l, *}}{N_l + \varepsilon},
\end{aligned}
\end{equation}
where $\lambda \in [0,1)$ is the decay rate and $\varepsilon$ is a small constant for numerical stability.

\paragraph{Grid-Sampled Semantic Supervision.}
Grid sampling aims to reduce the resolution of the semantic map (height and width, i.e., the number of pixels). Applying dense semantic supervision at every image pixel is unnecessarily expensive during mapping, as each 3D Gaussian typically projects to a 2D splat covering many pixels. To reduce this cost, instead of rendering a full-resolution semantic feature map, we supervise the features on a sparse, regular grid defined by a specific stride and offset.

Crucially, rather than relying on standard high-level tensor subsampling, we implement this via \textbf{a custom GPU kernel written in CUDA}. This low-level optimization performs only the minimal pixel-level calculations required for the subsampled locations. By doing so, it entirely avoids the computational and memory burdens of handling a full-resolution semantic supervision map.

During optimization, we compute a combined L1 and cosine similarity loss between the rendered sparse features and the ground-truth features. We pair this with a binary validity mask to filter out unannotated background regions (preventing Gaussians from learning empty features) and to handle out-of-bounds coordinates. With stride $s$, our rasterizer reduces supervised pixels, memory use, and computation by $s^2$ during online mapping.

\paragraph{Parallelized Pipeline.}
To integrate these individual components into a highly efficient workflow, we employ a pipeline with highly parallelized scheduling, as illustrated in Figure~\ref{fig:overview}. The VQ codebook update is executed as soon as the VFM completes encoding an incoming keyframe. Simultaneously, we initiate a ``grid-sampled target prefetching'' operation on the semantic map. Because operations across different grid offsets are mutually independent, we allocate multiple background workers to extract and buffer the grid-sampled targets well in advance of the actual optimization loop.

Crucially, to maintain system stability and reduce peak computational load, we decouple the geometry and appearance updates from the semantic updates. During the semantic distillation phase, all other parameters of the 3D Gaussians (i.e., position, scale, rotation, opacity, and color) remain frozen. Conversely, the learned semantic logits and VQ codebooks are strictly excluded from the base tracking and mapping phases. This design reduces distillation costs and supports real-time rendering.

\subsection{X-GS-\textit{Thinker}}
X-GS-\textit{Thinker} interfaces the resulting semantic Gaussians with diverse VLM architectures, enabling spatial multimodal capabilities.

\subsubsection{Contrastive VLM}
For 3D visual grounding, the \textit{Thinker} employs contrastive VLMs consisting of an image encoder and a text encoder, such as CLIP \cite{DBLP:conf/icml/RadfordKHRGASAM21}, as well as VFMs for segmentation \cite{DBLP:journals/corr/abs-2304-02643}. Since the 3D Gaussians from the \textit{Perceiver} already encapsulate distilled image embeddings as semantic features, we can directly query and identify target objects by computing the similarity between the Gaussians' semantic features, decoded via the online VQ module, and the prompt embeddings produced by the text encoder.

\input{tables/2_ground}

\subsubsection{Generative VLM}
To achieve 3D scene understanding, the \textit{Perceiver} distills the embeddings from the vision encoder of a generative VLM, ensuring these semantic features can be directly utilized by the VLM's backbone, acting as the \textit{Thinker}. Because utilizing the full set of $N_g$ Gaussians is computationally prohibitive and yields highly redundant tokens, we condense the scene into a compact sequence of $M \ll N_g$ informative tokens using an Entropy-Adaptive Gaussian Sampling strategy \cite{DBLP:journals/corr/abs-2503-06271}. Since the underlying 3DGS map is constructed from sequential camera observations, this approach effectively allows X-GS to act as a powerful compression mechanism, distilling long video sequences into a concise set of informative tokens for the VLM.

%% file: figures/2_deps_fig.tex
\begin{figure*}[t]
\centering
\includegraphics[width=1.0\textwidth, trim={0.25in 11.935in 0.75in 0.875in},clip]{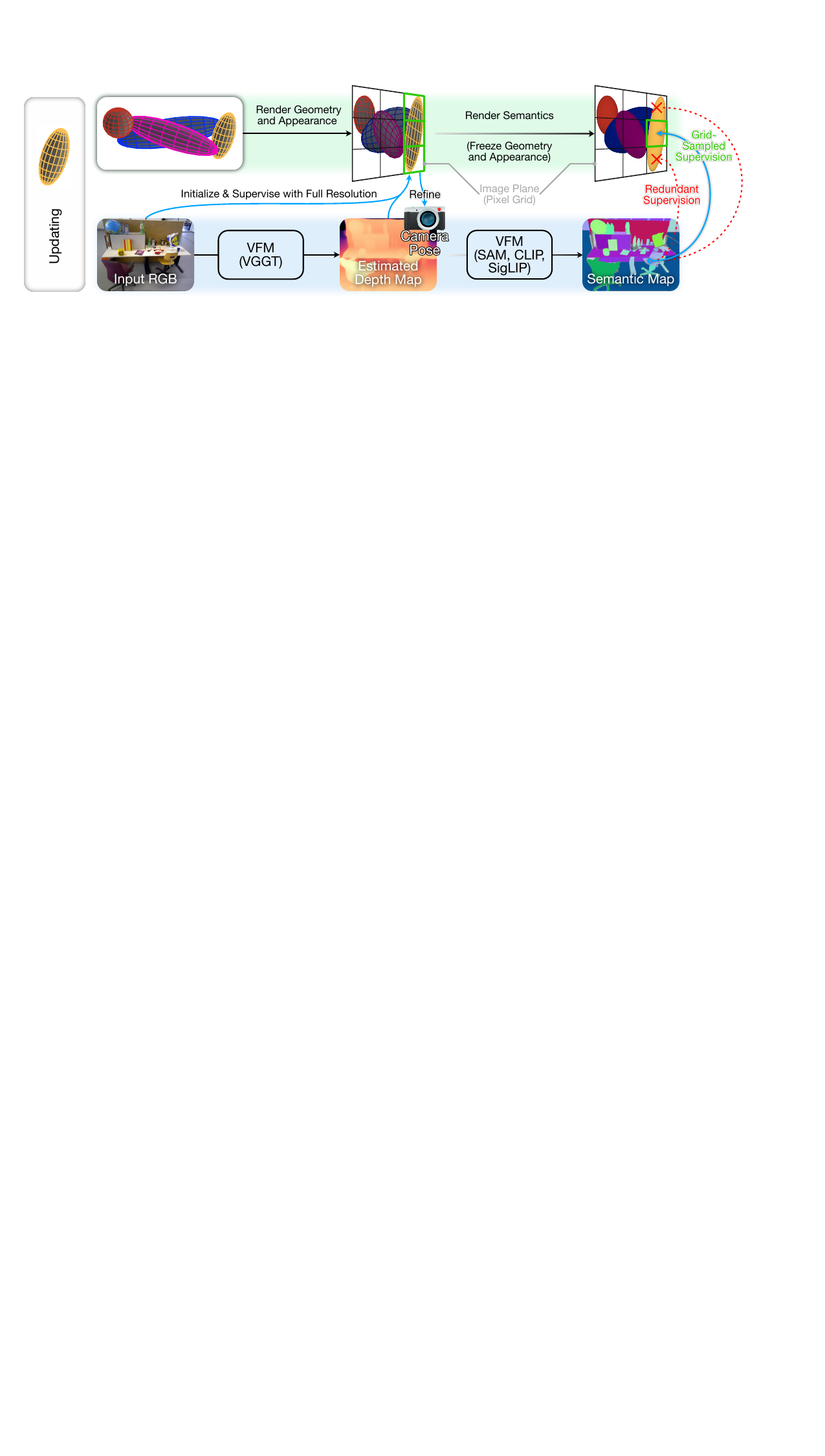}
\caption{Illustration of the information flow among the core components of X-GS-\textit{Perceiver}, alongside an illustrative example of grid-sampled semantic supervision.}
\label{fig:deps}
\end{figure*}

%% file: tables/1_slam.tex
\begin{table*}[t]
    \centering
    \newcommand{\g}[1]{\textcolor{gray}{#1}}
    \caption{\textbf{Quantitative evaluation on SLAM benchmarks.} RGB-D methods (\g{gray numbers}) utilize ground-truth depth and are therefore not directly comparable to monocular (RGB-only) baselines or to X-GS-\textit{Perceiver}. Rows highlighted in gray are ablation studies. ATE denotes absolute trajectory error, reported as RMSE (cm). PCD: point clouds. FF: feed-forward. Alt.: alternative depth prior (VGGT-SLAM in place of StreamVGGT).}
    \label{tab:slam_results}
    \resizebox{\textwidth}{!}{%
        \begin{tabular}{l l >{\Centering}c >{\Centering}c>{\Centering}c>{\Centering}c >{\Centering}c>{\Centering}c>{\Centering}c>{\Centering}c}
            \Xhline{1.2pt}
            \multicolumn{1}{l!{\vrule}}{\multirow{2}{*}{\textbf{Method}}} &
                \multicolumn{1}{l!{\vrule}}{\multirow{2}{*}{\textbf{Paradigm}}} &
                \multicolumn{1}{c!{\vrule}}{\textbf{Tracking}} &
                \multicolumn{3}{c!{\vrule}}{\textbf{Appearance (RGB)}} &
                \multicolumn{4}{c}{\textbf{Geometry (Depth)}} \\
            \cline{3-10}
            \multicolumn{1}{l!{\vrule}}{} &
                \multicolumn{1}{l!{\vrule}}{} &
                \multicolumn{1}{c!{\vrule}}{ATE \(\downarrow\)} &
                PSNR \(\uparrow\) &
                SSIM \(\uparrow\) &
                \multicolumn{1}{c!{\vrule}}{LPIPS \(\downarrow\)} &
                Abs Rel \(\downarrow\) &
                Sq Rel \(\downarrow\) &
                RMSE \(\downarrow\) &
                \(\delta < 1.25 \uparrow\) \\

            \specialrule{1.1pt}{0.2ex}{0.2ex}
            \multicolumn{10}{c}{\large\textsc{Benchmark}:~\textbf{TUM RGB-D}} \\[0.25ex]
            \multicolumn{10}{l}{\textit{\textbf{RGB-D}}} \\
            Photo-SLAM \cite{DBLP:conf/cvpr/HuangLCY24} & RGB-D SLAM & \g{1.80} & \g{19.28} & \g{0.673} & \g{0.334} & \g{\NA} & \g{\NA} & \g{\NA} & \g{\NA} \\
            SEGS-SLAM \cite{DBLP:conf/iccv/WenLF25} & RGB-D SLAM & \g{1.14} & \g{23.20} & \g{0.769} & \g{0.271} & \g{\NA} & \g{\NA} & \g{\NA} & \g{\NA} \\
            \hline
            \multicolumn{10}{l}{\textit{\textbf{Monocular}}} \\
            SplatTalk \cite{DBLP:journals/corr/abs-2503-06271} & Offline FF 3DGS & \NA              & 15.51            & 0.483            & 0.243            & 0.795            & 5.301            & 7.006             & 0.000            \\
            StreamVGGT  \cite{DBLP:journals/corr/abs-2507-11539} & Online PCD   & 2.70             & 13.58            & 0.281            & 0.594            & 0.105            & 0.076            & 0.500             & 0.879            \\
            MonoGS  \cite{DBLP:conf/cvpr/MatsukiMKD24} & Online 3DGS  & 3.96             & 21.89            & 0.733            & 0.327            & 0.352            & 0.292            & 0.666             & 0.446            \\
            \hline
            X-GS-\textit{Perceiver} (Ours) & Online 3DGS  & \textbf{1.55}    & \textbf{25.21}   & \textbf{0.784}   & \textbf{0.236}   & \textbf{0.055}   & \textbf{0.032}   & \textbf{0.326}    & \textbf{0.945}   \\
            \rowcolor{gray!15} \textit{\textbf{Ablation Studies}} & & & & & & & & & \\
            \rowcolor{gray!15} W/o StreamVGGT & No depth prior & 4.49 & 23.66 & 0.746 & 0.256 & 0.162 & 0.126 & 0.616 & 0.746 \\
            \rowcolor{gray!15} With VGGT-SLAM & Alt. depth prior & 3.11 & 23.42 & 0.754 & 0.252 & 0.063 & 0.047 & 0.387 & 0.938 \\

            \specialrule{0.5pt}{0.6ex}{0.45pt}
            \specialrule{0.5pt}{0.45pt}{0.35ex}
            \multicolumn{10}{c}{\large\textsc{Benchmark}:~\textbf{Replica}} \\[0.25ex]
            \multicolumn{10}{l}{\textit{\textbf{RGB-D}}} \\
            NICE-SLAM \cite{DBLP:conf/cvpr/ZhuPLXBCOP22} & RGB-D SLAM & \g{1.07} & \g{24.42} & \g{0.809} & \g{0.233} & \g{\NA} & \g{\NA} & \g{\NA} & \g{\NA} \\
            Vox-Fusion \cite{DBLP:conf/ismar/0001LZML022} & RGB-D SLAM & \g{3.09} & \g{24.41} & \g{0.801} & \g{0.236} & \g{\NA} & \g{\NA} & \g{\NA} & \g{\NA} \\
            Photo-SLAM & RGB-D SLAM & \g{10.63} & \g{30.37} & \g{0.904} & \g{0.161} & \g{\NA} & \g{\NA} & \g{\NA} & \g{\NA} \\
            SEGS-SLAM & RGB-D SLAM & \g{9.25} & \g{33.54} & \g{0.927} & \g{0.104} & \g{\NA} & \g{\NA} & \g{\NA} & \g{\NA} \\
            \hline
            \multicolumn{10}{l}{\textit{\textbf{Monocular}}} \\
            SplatTalk              & Offline FF 3DGS & \NA              & 16.90            & 0.677            & 0.431            & 0.807            & 11.228           & 14.416           & 0.006            \\
            StreamVGGT             & Online PCD   & 5.93            & 7.32             & 0.044            & 1.125            & 0.030            & 0.013            & 0.159            & 0.980            \\
            MonoGS                 & Online 3DGS  & 2.80             & 29.31            & 0.892            & 0.231            & 0.155            & 0.104            & 0.500            & 0.763            \\
            \hline
            X-GS-\textit{Perceiver} (Ours) & Online 3DGS  & 1.12             & \textbf{34.66}   & \textbf{0.937}   & \textbf{0.115}   & 0.022            & 0.008            & 0.094             & 0.991            \\
            \rowcolor{gray!15} \textit{\textbf{Ablation Studies}} & & & & & & & & & \\
            \rowcolor{gray!15} W/o StreamVGGT & No depth prior & 3.92 & 29.59 & 0.889 & 0.180 & 0.127 & 0.077 & 0.396 & 0.836 \\
            \rowcolor{gray!15} With VGGT-SLAM & Alt. depth prior & \textbf{0.81} & 33.00 & 0.926 & 0.132 & \textbf{0.016} & \textbf{0.003} & \textbf{0.076} & \textbf{0.996} \\

            \Xhline{1.2pt}
        \end{tabular}%
    }
\end{table*}

%% file: figures/3_qualitative_fig.tex
\begin{figure*}[t]
\centering
\includegraphics[width=1.0\textwidth, trim={0.125in 1.55in 3in 0.35in},clip]{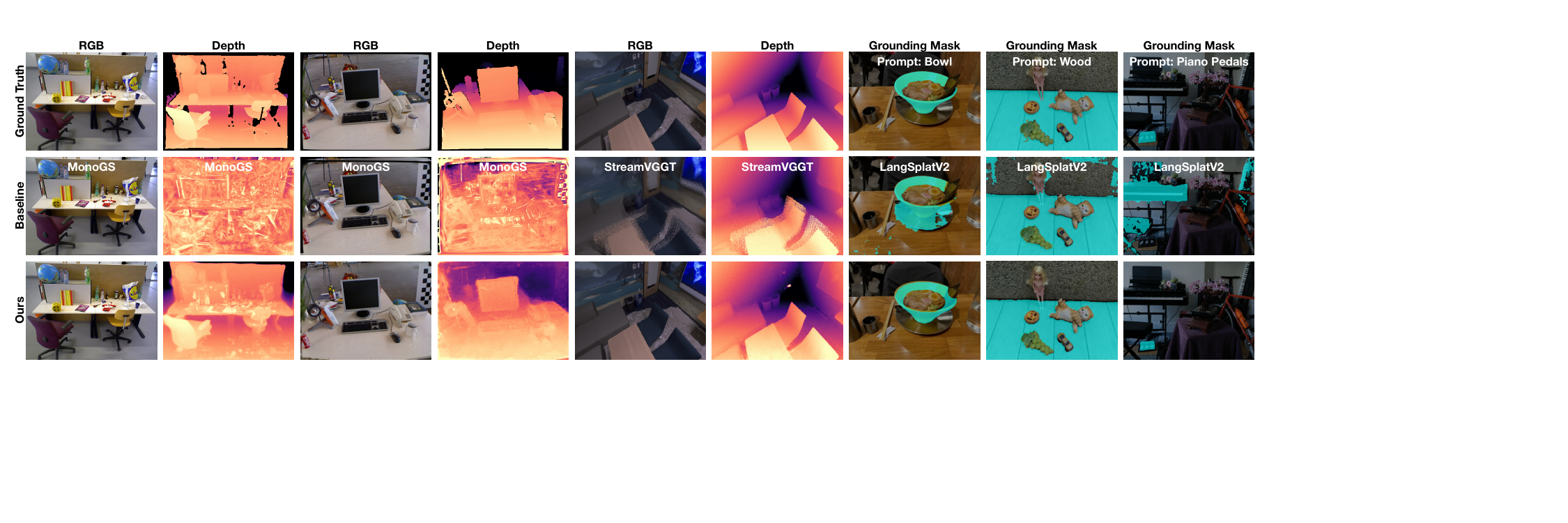}
\caption{\textbf{Qualitative Comparisons.} Visual comparisons of X-GS against baseline methods and ground truth (GT) on scene reconstruction and 3D visual grounding tasks.}
\label{fig:qualitative}
\end{figure*}

%% file: tables/2_ground.tex
\begin{table}[t]
    \caption{\textbf{Quantitative results on 3D visual grounding benchmarks.} Contrastive X-GS is reported using overall mAcc for LERF Localization and overall mIoU for LERF Segmentation and 3D-OVS. Times in parentheses, taken from Table~\ref{tab:time} (CTR), show default $\rightarrow$ ablation runtimes.}
    \label{tab:grounding}
    \centering
    \fontsize{7.5pt}{9.5pt}\selectfont
    \setlength{\tabcolsep}{3pt}
    \begin{tabularx}{\columnwidth}{l *{3}{>{\centering\arraybackslash}X}}
        \Xhline{0.8pt}
        \multicolumn{1}{l!{\vrule}}{\multirow{2}{*}{\textbf{Method}}} &
            \multicolumn{2}{c!{\vrule}}{\textbf{LERF}} &
            \textbf{3D-OVS} \\
        \cline{2-4}
        \multicolumn{1}{l!{\vrule}}{} &
            mAcc &
            \multicolumn{1}{c!{\vrule}}{mIoU} &
            mIoU \\
        \hline

        FFD \cite{DBLP:conf/nips/KobayashiMS22} &
            \NA & \NA & 26.9 \\
        LSeg \cite{DBLP:conf/iclr/LiWBKR22} &
            21.1 & 16.6 & 20.6 \\
        LERF \cite{DBLP:conf/iccv/KerrKGKT23} &
            73.6 & 37.4 & 54.8 \\
        3D-OVS \cite{DBLP:conf/nips/LiuZZXYETXL23} &
            \NA & \NA & 86.8 \\
        GS-Grouping \cite{DBLP:conf/eccv/YeDYK24} &
            49.1 & 36.6 & 87.7 \\
        LEGaussian \cite{DBLP:conf/cvpr/ShiWDG24} &
            67.4 & 24.6 & 88.5 \\
        GOI \cite{DBLP:conf/mm/QuDLLCZJ24} &
            59.2 & 42.0 & 90.6 \\
        GAGS \cite{peng2024gags} &
            81.7 & 54.1 & \NA \\
        LangSplat \cite{DBLP:conf/cvpr/Qin0ZWP24} &
            84.3 & 51.4 & 93.4 \\
        LangSplatV2 \cite{DBLP:journals/corr/abs-2507-07136} &
            84.1 & 59.9 & 94.6 \\

        \hline
        X-GS & 89.7 & 61.8 & 94.3 \\

        \rowcolor{gray!15} \textit{\textbf{Ablation Studies}} & & & \\
        \rowcolor{gray!15} W/o StreamVGGT & 82.8 & 56.6 & 93.5 \\
        \rowcolor{gray!15} \hspace*{1em}{\tiny(With RGB only: 3.04s $\rightarrow$ 0s)} & & & \\
        \rowcolor{gray!15} W/o MobileSAMv2 & \textbf{89.9} & 61.9 & \textbf{95.4} \\
        \rowcolor{gray!15} \hspace*{1em}{\tiny(With SAM: 3.87s $\rightarrow$ 73.73s)} & & & \\
        \rowcolor{gray!15} W/o grid sampling & 89.4 & \textbf{62.6} & 94.8 \\
        \rowcolor{gray!15} \hspace*{1em}{\tiny(With full resolution: 4.27s $\rightarrow$ 37.02s)} & & & \\
        \rowcolor{gray!15} W/o online distillation & 88.9 & 62.3 & 94.2 \\
        \rowcolor{gray!15} W/o VQ module & \multicolumn{3}{c}{\cellcolor{gray!15}(Out of Memory)} \\

        \Xhline{0.8pt}
    \end{tabularx}
\end{table}

%% file: sections/4_experiment.tex
\section{Experiments}
\input{tables/3_3dvl}
\input{tables/4_time}

\subsection{Implementation Details}
For the X-GS-\textit{Perceiver}, we utilize a weight of $0.05$ for the depth map estimated by StreamVGGT \cite{DBLP:journals/corr/abs-2507-11539}, and we scale the learning rate of the position parameters by a factor of $0.2$. The occluder pruning mechanism uses a relative margin of $10\%$. The VQ codebook is instantiated with $L = 64$ independent semantic codewords. During the online codebook learning stage, the EMA update incorporates a decay momentum of $\lambda = 0.96$ and a temperature of $\tau = 0.7$ to smoothly adapt the cluster centers to the sequential frame stream. We maintain a sliding window of $N_{win} = 8$ keyframes, allocating $150$ optimization iterations for the geometric and photometric mapping phase, followed by $50$ iterations dedicated solely to the semantic distillation phase. The \textit{Perceiver} can operate efficiently on a single NVIDIA V100 GPU.

\subsection{SLAM Benchmarks}

We evaluate X-GS-\textit{Perceiver} on two standard SLAM benchmarks, TUM RGB-D \cite{DBLP:conf/iros/SturmEEBC12} and Replica \cite{DBLP:journals/corr/abs-1906-05797}, assessing tracking accuracy, photometric rendering quality (RGB), and geometric reconstruction accuracy (depth). Except for the RGB-D SLAM baselines and SplatTalk, which uses posed RGB images, all evaluated models rely solely on monocular RGB inputs and operate without access to ground-truth depth, camera poses, or auxiliary data. Quantitative results are summarized in Table~\ref{tab:slam_results}.

By leveraging recent VFMs capable of simultaneous camera pose and depth estimation, X-GS-\textit{Perceiver} outperforms MonoGS in both tracking and rendering quality. Furthermore, supervising the 3D Gaussians with the estimated depth maps yields significant improvements in geometric accuracy. Qualitative comparisons are provided in Figure~\ref{fig:qualitative}.

\subsection{3D Visual Grounding Benchmarks}

Under the contrastive paradigm of X-GS, we prompt the \textit{Thinker} to identify relevant 3D Gaussians, effectively grounding natural language queries within the perceived 3D scene. We evaluate our method on two 3D visual grounding benchmarks: LERF \cite{DBLP:conf/iccv/KerrKGKT23} and 3D-OVS \cite{DBLP:conf/nips/LiuZZXYETXL23}. Baselines and quantitative results are reported in Table~\ref{tab:grounding}.

Our results demonstrate that X-GS outperforms LangSplatV2 on LERF and matches it on 3D-OVS, which we attribute to the superior geometry from StreamVGGT. Qualitative comparisons are provided in Figure~\ref{fig:qualitative}. Furthermore, because LangSplatV2 relies on SAM 1.0 \cite{DBLP:journals/corr/abs-2304-02643} and requires full-resolution semantic maps for supervision, its training process is computationally expensive (see Table~\ref{tab:time}). By contrast, our accelerated online semantic distillation strategy, coupled with the integration of MobileSAMv2 \cite{DBLP:journals/corr/abs-2312-09579}, dramatically increases processing speed with negligible loss in visual grounding quality.

\subsection{3D Vision-Language Benchmarks}

To evaluate the 3D understanding and reasoning capabilities of X-GS, we employ three representative 3D vision-language (3D-VL) benchmarks: Scan2Cap \cite{DBLP:conf/cvpr/ChenGNC21}, ScanQA \cite{DBLP:conf/cvpr/AzumaMKK22}, and SQA3D \cite{DBLP:conf/iclr/MaYZ0LZH23}. Quantitative results are presented in Table~\ref{tab:3dvl}.

We instantiate the X-GS-\textit{Thinker} using LLaVA-OneVision~\cite{DBLP:journals/tmlr/0080ZGZ00ZZL0L25}, backed by the Qwen2-7B-Instruct LLM~\cite{DBLP:journals/corr/abs-2407-10671}. In this configuration, the \textit{Thinker} interfaces the VLM with semantic Gaussians: it directly ingests 3D Gaussians containing semantic features distilled from the SigLIP \cite{DBLP:conf/iccv/ZhaiM0B23} VFM, which is the default vision encoder for LLaVA-OneVision. By acting as an efficient compression mechanism for these semantic features, X-GS achieves improved zero-shot performance compared to the base LLaVA-OneVision model. Furthermore, following LoRA \cite{DBLP:conf/iclr/HuSWALWWC22} finetuning, X-GS outperforms specialized 3D VLMs that are natively designed for these 3D-VL tasks.

\subsection{Ablation Studies}
We ablate the key components of X-GS in the corresponding benchmark tables (\colorbox{gray!15}{gray rows}). On SLAM (Table~\ref{tab:slam_results}), removing StreamVGGT isolates the geometry prior. For 3D visual grounding (Table~\ref{tab:grounding}), we ablate StreamVGGT, MobileSAMv2, grid sampling, the VQ module, and online semantic distillation. Table~\ref{tab:3dvl} ablates the zero-shot \textit{Thinker}. Replacing reconstructed Gaussians with 2D tokens lowers accuracy across benchmarks, while entropy-adaptive sampling from the 3D map remains comparatively strong and outperforms spatially uniform sampling, indicating that the spatial structure of 3DGS is beneficial for 3D vision-language tasks. Runtime is analyzed in Section~\ref{sec:time}.

These results also provide experimental evidence of extensibility: alternative modules can be integrated into the same X-GS slots. VGGT-SLAM \cite{DBLP:journals/corr/abs-2505-12549} can be used in place of StreamVGGT at the \textit{Perceiver} pose/depth interface (Table~\ref{tab:slam_results}). Unlike StreamVGGT, VGGT-SLAM is submap-based rather than per-frame streaming. SAM~1.0 can be used instead of MobileSAMv2 for semantic segmentation (Table~\ref{tab:grounding}). CLIP and SigLIP are both compatible with X-GS's semantic distillation, supporting the contrastive \textit{Thinker} in Table~\ref{tab:grounding} and the generative \textit{Thinker} in Table~\ref{tab:3dvl}, respectively.

\subsection{Computational Footprint Analysis}
\label{sec:time}

Table~\ref{tab:time} breaks down the runtime of the contrastive (CTR) and generative (GEN) variants of X-GS-\textit{Perceiver} in single-thread (ST) and multi-thread (MT) settings \cite{DBLP:conf/cvpr/MatsukiMKD24}. The resulting semantic Gaussians render in real time: contrastive X-GS reaches $312.5$ RGB FPS and $105.4$ semantic FPS, versus $304.9$ and $85.6$ for LangSplatV2. Our efficiency gains concentrate on semantic distillation. For CTR, replacing SAM 1.0 \cite{DBLP:journals/corr/abs-2304-02643} with MobileSAMv2 \cite{DBLP:journals/corr/abs-2312-09579} cuts vision encoding from $73.73$ s to $3.87$ s per keyframe. Multi-worker prefetching cuts target construction from $25.87$ s to $0.40$ s (CTR) and $1.34$ s (GEN). With grid-sampled supervision, the semantic forward and backward passes take $4.27$ s (CTR) and $5.52$ s (GEN) in the MT setting, compared with $37.02$ s for LangSplatV2's full-resolution passes.

The parallelized pipeline further overlaps supervision preparation (vision encoding, the EMA codebook update, and target construction) with geometry and appearance optimization. Overall, a full X-GS keyframe takes $10.22$ s (CTR) and $11.93$ s (GEN) in the MT setting, whereas LangSplatV2 alone spends $136.68$ s per keyframe on semantic distillation. Compared with MonoGS, which has no semantic channel ($2.41$ s per keyframe and $1.93$ FPS), X-GS runs at $0.41$ FPS (CTR) and $1.19$ FPS (GEN, whose ScanNet sequences have sparser keyframes). We consider this a moderate price for online semantic distillation.

%% file: tables/3_3dvl.tex
\begin{table*}[t]
    \centering
    \caption{\textbf{Quantitative results on 3D vision-language benchmarks.} The abbreviations C, B-4, M, R, and EM@1 denote CIDEr, BLEU-4, METEOR, ROUGE, and Exact Match, respectively.}
    \resizebox{1.0\textwidth}{!}{%
    \begin{tabular}{l
                    >{\Centering}p{1.2cm}
                    >{\Centering}p{1.2cm}
                    >{\Centering}p{1.2cm}
                    >{\Centering}p{1.2cm}
                    >{\Centering}p{1.2cm}
                    >{\Centering}p{1.2cm}
                    >{\Centering}p{1.2cm}
                    >{\Centering}p{1.2cm}
                    >{\Centering}p{1.2cm}
                    >{\Centering}p{1.2cm}
                    }
    \Xhline{1.2pt}

    \multicolumn{1}{l!{\vrule}}{\multirow{2}{*}{\textbf{Method}}} &
        \multicolumn{4}{c!{\vrule}}{\textbf{Scan2Cap}} &
        \multicolumn{5}{c!{\vrule}}{\textbf{ScanQA}} &
        \multicolumn{1}{c}{\textbf{SQA3D}}\\
    \cline{2-11}
        \multicolumn{1}{l!{\vrule}}{} &
        C &
        B-4 &
        M &
        \multicolumn{1}{c!{\vrule}}{R} &
        C &
        B-4 &
        M &
        R &
        \multicolumn{1}{c!{\vrule}}{EM@1} &
        EM@1 \\
        \hline

        \textit{\textbf{Task-Specific Models}} & & & & & & & & & & \\
        Scan2Cap (GPT-3) \cite{DBLP:conf/cvpr/ChenGNC21} & 35.2 & 22.4 & 21.4 & 43.5 & \NA & \NA & \NA & \NA & \NA & 41.0 \\
        3DJCG \cite{DBLP:conf/cvpr/CaiZZSX22} & 47.7 & 31.5 & 24.3 & 51.8 & \NA & \NA & \NA & \NA & \NA & \NA \\
        Vote2Cap-DETR \cite{DBLP:conf/cvpr/ChenZCLYC23} & 61.8 & 34.5 & 26.2 & 54.4 & \NA & \NA & \NA & \NA & \NA & \NA \\
        ScanRefer+MCAN \cite{DBLP:conf/eccv/ChenCN20} & \NA & \NA & \NA & \NA & 55.4 & 7.9 & 11.5 & 30.0 & 18.6 & \NA \\
        ClipBERT \cite{DBLP:conf/cvpr/LeiLZGBB021} & \NA & \NA & \NA & \NA & \NA & \NA & \NA & \NA & \NA & 43.3 \\
        ScanQA \cite{DBLP:conf/cvpr/AzumaMKK22} & \NA & \NA & \NA & \NA & 64.9 & 10.1 & 13.1 & 33.3 & 21.1 & 47.2 \\
            \hline

        \textit{\textbf{3D VLMs}} & & & & & & & & & & \\
        3D-VisTA \cite{DBLP:conf/iccv/ZhuMCD0023} &
            66.9  &  34.0  &  27.1  &  54.3  &  69.6  &  10.4  &  13.9  &  35.7  &  22.4  &  48.5
            \\
        3D-LLM \cite{DBLP:conf/nips/HongZCZDCG23} &
            \NA  &  8.1  &  13.1  &  33.2  &  69.4  &  12.0  &  14.5  &  35.7  &  \NA  &  \NA
            \\
        LL3DA (Task-Specific) \cite{DBLP:conf/cvpr/ChenCZLY0Z0024} &
            65.2  &  36.8  &  26.0  &  55.1  &  76.8  &  13.5  &  15.9  &  37.3  &  \NA  &  \NA
            \\
        LL3DA (Generalist) &
            63.0  &  36.0  &  25.7  &  54.7  &  75.7  &  13.3  &  15.4  &  37.0  &  \NA  &  \NA
            \\
        Scene-LLM (Task-Specific) \cite{DBLP:journals/corr/abs-2403-11401} &
            37.9  &  24.1  &  21.8  &  45.6  &  80.0  &  12.0  &  16.6  &  40.0  &  \NA  &  54.2
            \\
        Scene-LLM (Generalist) &
            \NA  &  \NA  &  \NA  &  \NA  &  80.0  &  11.7  &  15.8  &  35.9  &  \NA  &  53.6
            \\
            \hline

        \textit{\textbf{2D VLMs}} & & & & & & & & & & \\
        MoE-LLaVA \cite{DBLP:journals/corr/abs-2401-15947} &
            27.4  &  18.1  &  12.1  &  40.5  &  42.5  &  5.0  &  7.7  &  21.8  &  11.3  &  22.3
            \\
        LLaVA-NeXT \cite{liu2024llavanext} & 
            36.0  &  20.4  &  14.2  &  49.5  &  48.6  &  6.3  &  9.8  &  22.3  &  14.6  &  23.5
            \\

        LLaVA-OneVision (Zero-Shot) \cite{DBLP:journals/tmlr/0080ZGZ00ZZL0L25} & 
            28.3  &  4.7  &  11.6  &  32.4  &  50.0  &  \NA  &  13.0  &  29.4  &  15.6  &  25.4
            \\
            \hline

        \textit{\textbf{2D VLM-Based Models}} & & & & & & & & & & \\
        SplatTalk \cite{DBLP:journals/corr/abs-2503-06271} &
            \NA  &  \NA  &  \NA  &  \NA  &  61.7  &  \NA  &  14.2  &  32.7  &  17.1  &  26.1
            \\
        SplatTalk-ScanQA-FT &
            \NA  &  \NA  &  \NA  &  \NA  &  77.1  &  \NA  &  15.4  &  38.7  &  22.3  &  42.0
            \\
        SplatTalk-3DVQA-FT &
            \NA  &  \NA  &  \NA  &  \NA  &  77.5  &  \NA  &  15.6  &  38.5  &  22.4  &  47.6
            \\

            \hline

        X-GS (Zero-Shot) &
            31.8  &  5.9  &  12.3  &  38.5  &  64.0  &  7.1  &  15.1  &  33.4  &  19.3  &  27.0
            \\
        X-GS (Finetuned) &
            \textbf{68.3}  &  \textbf{38.3}  &  \textbf{27.5}  &  \textbf{55.5}  &  \textbf{83.8}  &  \textbf{15.3}  &  \textbf{17.2}  &  \textbf{41.5}  &  \textbf{23.1}  &  \textbf{54.9}
            \\

        \rowcolor{gray!15} \textit{\textbf{Ablation Studies (Zero-Shot)}} & & & & & & & & & & \\
        \rowcolor{gray!15} W/o X-GS-\textit{Perceiver} &
            27.5  &  4.5  &  11.3  &  31.9  &  49.2  &  5.0  &  12.9  &  29.2  &  15.1  &  24.6 \\
        \rowcolor{gray!15} W/o entropy-adaptive sampling (use all tokens) &
            19.1  &  3.1  &  7.9  &  22.7  &  45.4  &  4.4  &  11.8  &  27.2  &  9.9  &  22.8 \\
        \rowcolor{gray!15} With spatially uniform sampling &
            28.2  &  5.0  &  11.4  &  34.6  &  60.4  &  6.6  &  14.0  &  31.5  &  18.1  &  25.3 \\

        \Xhline{1.2pt}
    \end{tabular}
    }
    \label{tab:3dvl}
\end{table*}

%% file: tables/4_time.tex
\begin{table*}[t]
    \centering
    \caption{\textbf{Computational footprint analysis of X-GS-\textit{Perceiver}, run on a single NVIDIA V100 GPU.} For X-GS, we report the two variants separately: contrastive VLM (CTR), evaluated on TUM RGB-D, and generative VLM (GEN), evaluated on ScanNet. Paired times are for single-thread (ST) and multi-thread (MT) settings.
    }
    \label{tab:time}
    \scriptsize
    \newdimen\xgstw
    \settowidth{\xgstw}{39.78}
    \newdimen\xgsbarh
    \newdimen\xgsbard
    \settoheight{\xgsbarh}{|}
    \settodepth{\xgsbard}{|}
    \addtolength{\xgsbarh}{0.12ex}
    \addtolength{\xgsbard}{0.12ex}
    \newcommand{\xgsbar}{\hspace{0.25em}\vrule width 0.3pt height\xgsbarh depth\xgsbard\hspace{0.25em}}
    \setlength{\tabcolsep}{4pt}
    \resizebox{\linewidth}{!}{%
    \begin{tabular}{
            l
            r w{r}{\xgstw} @{\xgsbar} w{l}{\xgstw}
            r w{r}{\xgstw} @{\xgsbar} w{l}{\xgstw}
            r w{r}{\xgstw} @{\xgsbar} w{l}{\xgstw}
        }
        \Xhline{0.9pt}
        \multicolumn{1}{l!{\vrule}}{\multirow{2}{*}{\textbf{Main Component}}} &
            \multicolumn{3}{c!{\vrule}}{\textbf{Baseline}} &
            \multicolumn{3}{c!{\vrule}}{\textbf{X-GS} (CTR)} &
            \multicolumn{3}{c}{\textbf{X-GS} (GEN)} \\
        \cline{2-10}
        \multicolumn{1}{l!{\vrule}}{} &
            \multicolumn{1}{c}{\textbf{Module}} &
            \multicolumn{2}{c!{\vrule}}{\makecell{\textbf{Time}\\[-0.4ex]{\normalfont\scriptsize ST\xgsbar MT}}} &
            \multicolumn{1}{c}{\textbf{Module}} &
            \multicolumn{2}{c!{\vrule}}{\makecell{\textbf{Time}\\[-0.4ex]{\normalfont\scriptsize ST\xgsbar MT}}} &
            \multicolumn{1}{c}{\textbf{Module}} &
            \multicolumn{2}{c}{\makecell{\textbf{Time}\\[-0.4ex]{\normalfont\scriptsize ST\xgsbar MT}}} \\
        \hline
        \textit{\textbf{Geometry \& Appearance Opt.}} &
            \multicolumn{3}{l}{(Baseline: MonoGS)} & \multicolumn{3}{c}{} & \multicolumn{3}{c}{} \\
        VFM Depth Est. & \NA & \multicolumn{2}{c}{\NA} & Online VGGT & \multicolumn{2}{c}{3.04} & Online VGGT & \multicolumn{2}{c}{0.78} \\
        Forward Pass & RGB & 10.40 & 0.76 & RGB-D & 11.52 & 0.85 & RGB-D & 8.76 & 0.65 \\
        Backward Pass & RGB & 2.77 & 0.20 & RGB-D & 3.09 & 0.23 & RGB-D & 4.25 & 0.30 \\
        \hline
        \textit{\textbf{Semantic Supervision Prep.}} &
            \multicolumn{3}{l}{(Baseline: LangSplatV2)} & \multicolumn{3}{c}{} & \multicolumn{3}{c}{} \\
        Vision Encoding & SAM + CLIP & \multicolumn{2}{c}{73.73} & MobileSAMv2+CLIP & \multicolumn{2}{c}{3.87} & SigLIP+Proj. & \multicolumn{2}{c}{0.11} \\
        VQ Codebook & Offline & \multicolumn{2}{c}{0.06} & Online EMA & \multicolumn{2}{c}{0.04} & Online EMA & \multicolumn{2}{c}{0.91} \\
        Target Construction & Offline & \multicolumn{2}{c}{25.87} & Multi-Worker Prefetch & \multicolumn{2}{c}{0.40} & Multi-Worker Prefetch & \multicolumn{2}{c}{1.34} \\
        \hline
        \textit{\textbf{Semantic Optimization}} &
            \multicolumn{3}{l}{(Baseline: LangSplatV2)} & \multicolumn{3}{c}{} & \multicolumn{3}{c}{} \\
        Forward Pass & Full-Resolution & \multicolumn{2}{c}{25.41} & Grid-Sampled & 3.37 & 0.71 & Grid-Sampled & 4.19 & 0.93 \\
        Backward Pass & Full-Resolution & \multicolumn{2}{c}{11.61} & Grid-Sampled & 16.80 & 3.56 & Grid-Sampled & 21.80 & 4.59 \\
        \hline
        \textit{\textbf{Overall}} &
            \multicolumn{3}{l}{(Baseline: MonoGS)} & \multicolumn{3}{c}{} & \multicolumn{3}{c}{} \\
        Keyframe Interval & & \multicolumn{2}{c}{4.63} & & \multicolumn{2}{c}{4.17} & & \multicolumn{2}{c}{14.20} \\
        Asynchronous Overlap & & 0.00 & 0.00 & & 4.31 & 1.08 & & 2.36 & 0.95 \\
        Time per Keyframe & & 13.17 & 2.41 & & 37.82 & 10.22 & & 39.78 & 11.93 \\
        Time per Frame & & 2.84 & 0.52 & & 9.07 & 2.45 & & 2.80 & 0.84 \\
        Execution FPS & & 0.35 & 1.93 & & 0.11 & 0.41 & & 0.36 & 1.19 \\
        \hline
        \textit{\textbf{Rendering}} &
            \multicolumn{3}{l}{(Baseline: LangSplatV2)} & \multicolumn{3}{c}{} & \multicolumn{3}{c}{} \\
        RGB FPS & & \multicolumn{2}{c}{304.90} & & \multicolumn{2}{c}{312.50} & & \multicolumn{2}{c}{\NA} \\
        Semantic FPS & & \multicolumn{2}{c}{85.60} & & \multicolumn{2}{c}{105.40} & & \multicolumn{2}{c}{\NA} \\
        \Xhline{0.9pt}
    \end{tabular}%
    }
\end{table*}

%% file: sections/5_appendix.tex
\section{3D Gaussian Splatting}
\label{sec:appendix_3dgs}

3D Gaussian Splatting (3DGS) represents a 3D scene using a collection of anisotropic 3D Gaussians, operating as a point-based alternative to continuous neural implicit representations. Each Gaussian $i$ in the comprehensive map parameter set $\Theta$ is explicitly defined by the following attributes:

\begin{itemize}
    \item \textbf{Mean vector (Position):} $\boldsymbol{\mu}_i \in \mathbb{R}^3$.
    \item \textbf{Covariance matrix:} $\boldsymbol{\Sigma}_i \in \mathbb{R}^{3 \times 3}$.
    \item \textbf{Opacity:} $\alpha_i \in [0, 1]$.
    \item \textbf{Color features:} $\mathbf{c}_i$ (normally represented as spherical harmonics for view-dependence, though often simplified to base RGB values in dense SLAM configurations).
\end{itemize}

The unnormalized spatial distribution of a 3D Gaussian evaluated at a spatial point $\mathbf{x} \in \mathbb{R}^3$ is defined as:
\begin{equation}
G(\mathbf{x}) = \exp\left(-\frac{1}{2}(\mathbf{x} - \boldsymbol{\mu}_i)^\top \boldsymbol{\Sigma}_i^{-1} (\mathbf{x} - \boldsymbol{\mu}_i)\right).
\end{equation}

To maintain positive semi-definiteness during gradient-based optimization, the covariance matrix $\boldsymbol{\Sigma}_i$ is parameterized by a scaling matrix $\mathbf{S}_i$ and a rotation matrix $\mathbf{R}_i$:
\begin{equation}
\boldsymbol{\Sigma}_i = \mathbf{R}_i \mathbf{S}_i \mathbf{S}_i^\top \mathbf{R}_i^\top.
\end{equation}

In order to render these primitives onto the image plane from a given camera pose, each 3D Gaussian is projected into a 2D elliptical splat. Given a viewing transformation $\mathbf{W}$ and the Jacobian $\mathbf{J}$ of the affine approximation of the projective transformation, the resulting 2D covariance matrix $\boldsymbol{\Sigma}'_i$ is calculated as:
\begin{equation}
\boldsymbol{\Sigma}'_i = \mathbf{J} \mathbf{W} \boldsymbol{\Sigma}_i \mathbf{W}^\top \mathbf{J}^\top.
\end{equation}

This continuous state update sets a framework perfectly suited for dense tracking and mapping.

\section{3DGS-based SLAM Objectives}
\label{sec:appendix_slam}
Continuous optical depth models compute the expected color $C$ along a camera ray by integrating transmittance $T(t)$, volume density $\sigma(t)$, and color $\mathbf{c}(t)$. 3DGS discretizes this continuous integral using a point-based approximation. For a set of $N_g$ depth-sorted 2D splats obtained by projecting 3D Gaussians onto the image plane given a camera pose $\mathbf{T} \in \text{SE}(3)$, the color at pixel $x$ evaluates to:
\begin{equation}\label{eq:discrete_render}
C(x \mid \mathbf{T}, \Theta) = \sum_{i=1}^{N_g} \mathbf{c}_i \alpha'_i \prod_{j=1}^{i-1} (1 - \alpha'_j),
\end{equation}
where $\alpha'_i$ is the 2D splat opacity. Depth $D(x)$ is similarly rendered by substituting color $\mathbf{c}_i$ with the ray depth $d_i$.

Camera tracking is performed via Analysis-by-Synthesis. For an incoming frame at timestamp $t$, the pose is initialized using the online VFM's camera pose estimate. Because the 3DGS rasterizer is natively differentiable, precise tracking is achieved by freezing the map parameters $\Theta$ and running gradient descent on the camera extrinsic parameters to minimize the photometric and depth losses (if depth inputs are available):
\begin{equation}\label{eq:loss_track}
\mathcal{L}_{\text{track}} = \mathcal{L}_{\text{photo}}(I_t, \hat{I}_t) + \lambda_d \| D_t - \hat{D}_t \|_1.
\end{equation}
The global mapping thread optimizes the scene parameters $\Theta$ over a sliding window of keyframes $\mathcal{K}$ using the same combined loss, augmented with an isotropic regularization term $\mathcal{L}_{\text{iso}}$ to prevent heavily elongated artifacts:
\begin{equation}\label{eq:loss_map_expanded}
\begin{aligned}
\mathcal{L}_{\text{map}} &= \sum_{k \in \mathcal{K}} \Big( \mathcal{L}_{\text{photo}}(I_k, \hat{I}_k) + \lambda_d \| D_k - \hat{D}_k \|_1 \Big) \\
&\quad + \lambda_{\text{iso}} \mathcal{L}_{\text{iso}}.
\end{aligned}
\end{equation}

\section{Online VQ Codebook Details}
\label{sec:appendix_vq}
The online VQ module compresses semantic features into a compact latent space, akin to autoencoders \cite{DBLP:conf/naacl/MaCLSZK24}. To stabilize the online EMA updates introduced in the main text, we implement active dead-code management and a warm-start initialization process.

Instead of starting from uniform counts, we initialize the EMA statistics using a confidence-filtered assignment pass over an initial buffer of semantic features. We transform the observed feature distances into confidence scores using a temperature parameter $\tau$:
\begin{equation}\label{eq:confidence_score}
\begin{aligned}
p_{nl} = \frac{\exp( -\|\mathbf{x}_n - \mathbf{E}_{l, *}\|_2 / \tau )}{\sum_{j=1}^{L} \exp( -\|\mathbf{x}_n - \mathbf{E}_{j, *}\|_2 / \tau )}.
\end{aligned}
\end{equation}
Only samples for which the maximum confidence exceeds a threshold $\gamma$ are allowed to contribute to the initialization of the counts $N_l$ and sums $\mathbf{M}_{l, *}$:
\begin{equation}\label{eq:mask_confidence}
m_n = \mathbb{I}\!\left[\max_l p_{nl} \ge \gamma\right].
\end{equation}
During the continuous online tracking stage, the confidence score masks unreliable incoming samples, while the actual codeword assignments use hard nearest-neighbor matching to ensure discrete latent separation.

Furthermore, to handle dead codes escaping the tracked field of view, we explicitly monitor the EMA accumulated mass. If a codeword's utilization falls below a dead-code threshold ($N_l < \delta$), it is immediately reinitialized:
\begin{equation}\label{eq:dead_code}
\begin{aligned}
\mathbf{M}_{l, *} \leftarrow \tilde{\mathbf{x}}, \qquad
N_l \leftarrow 1 + \varepsilon,
\end{aligned}
\end{equation}
where $\tilde{\mathbf{x}}$ is sampled directly from a small reservoir of recently observed semantic features.

\section{Grid-Sampled Semantic Supervision Details}
\label{sec:appendix_grid_sampling}

This section details the mathematical formulation of our grid-sampled semantic supervision and the extraction of its targets.

\paragraph{Dynamic Grid Sampling.}
Let the full image domain be \(\Omega = \{0,\dots,H-1\} \times \{0,\dots,W-1\}\). For a given stride \(s \ge 1\) and spatial offset \(\mathbf{o} = (o_h, o_w)\), we sample a sparse grid of coordinates:
\begin{equation}\label{eq:sampled_loc}
(u,v) = (o_h + ms,\; o_w + ns),
\end{equation}
for integer indices \(m,n\), yielding a compact spatial resolution of \(H_s \times W_s\). To prevent the optimized point cloud from overfitting to a static lattice, the offset \(\mathbf{o}\) is sampled uniformly from \(\{0,\dots,s-1\}^2\) at each optimization step. This ensures the sampled grid smoothly traverses all \(s^2\) possible pixel phases over time.

\paragraph{Compact Predictions and Targets.}
Rather than forming a dense feature map, our custom rasterizer directly computes the compact prediction \(\hat{\mathbf{G}} \in \mathbb{R}^{D \times H_s \times W_s}\), where \(\hat{\mathbf{G}}_{:,m,n}\) corresponds to the rendered feature at \((u,v)\). We construct a corresponding target map \(\mathbf{G}^\star\) and binary validity mask \(\mathbf{V} \in \{0,1\}^{1 \times H_s \times W_s}\) on this same grid, populated via one of two pathways depending on the semantic source:

\textbf{1. Precomputed Region-Indexed Annotations:} Given a discrete segmentation mask \(S\) (where \(-1\) denotes background) and a region-feature table \(\Phi \in \mathbb{R}^{R \times D}\), we query the table at each sampled location \((u,v)\):
\begin{equation}\label{eq:target_discrete}
\begin{aligned}
V_{m,n} &= \mathbb{I}[S(u,v) \neq -1], \\
\mathbf{G}^\star_{:,m,n} &=
\begin{cases}
\Phi_{S(u,v)}, & S(u,v) \neq -1, \\
\mathbf{0}, & S(u,v) = -1.
\end{cases}
\end{aligned}
\end{equation}

\textbf{2. Online Continuous Feature Maps:} For dense features streaming from an online VLM at a lower abstract resolution \(\mathbf{P} \in \mathbb{R}^{D \times H_f \times W_f}\), we dynamically align the image coordinates to the low-resolution grid:
\begin{equation}\label{eq:target_continuous}
\begin{aligned}
\tilde{u}_{m,n} &= \operatorname{round}\!\left( u \cdot \frac{H_f-1}{\max(H-1,1)} \right), \\
\tilde{v}_{m,n} &= \operatorname{round}\!\left( v \cdot \frac{W_f-1}{\max(W-1,1)} \right).
\end{aligned}
\end{equation}
This yields the target \(\mathbf{G}^\star_{:,m,n} = \mathbf{P}_{:,\tilde{u}_{m,n},\tilde{v}_{m,n}}\) with an unconstrained mask \(V_{m,n}=1\).

\paragraph{Masked Loss Evaluation.}
Semantic supervision is applied exclusively at valid sampled locations using a combination of cosine similarity and L1 distance:
\begin{equation}\label{eq:loss}
\begin{aligned}
\mathcal{L}_{\mathrm{grid}} =\; &\lambda_{\mathrm{sem}} \left[ \mathcal{L}_{\cos}\!\left(\mathbf{V}\odot\hat{\mathbf{G}},\; \mathbf{V}\odot\mathbf{G}^\star\right) \right. \\
&\quad \left. + \, \mathcal{L}_{1}\!\left(\mathbf{V}\odot\hat{\mathbf{G}},\; \mathbf{V}\odot\mathbf{G}^\star\right) \right],
\end{aligned}
\end{equation}
where \(\odot\) denotes element-wise masking. The mask \(\mathbf{V}\) serves a dual purpose: it filters out unannotated background regions (preventing Gaussians from learning empty features) and masks out padded, out-of-bounds coordinates that arise from maintaining rigid batched tensor shapes at large grid offsets.

\input{tables/a1_benchmark.tex}

\section{3D Visual Grounding Details}
\label{sec:appendix_grounding}

Given a text prompt \(p\), we encode it with OpenCLIP and normalize the resulting embedding:
\begin{equation}
\mathbf{t}^{+}
=
\frac{E_{\text{text}}(p)}
{\left\|E_{\text{text}}(p)\right\|_2}.
\end{equation}

We also encode a small set of generic negative phrases
\begin{equation}
\mathcal{N}
=
\{\text{object},\ \text{things},\ \text{stuff},\ \text{texture}\},
\end{equation}
with normalized embeddings
\begin{equation}
\mathbf{t}_{n}^{-}
=
\frac{E_{\text{text}}(n)}
{\left\|E_{\text{text}}(n)\right\|_2},
\qquad n \in \mathcal{N}.
\end{equation}

As described in the main text, each Gaussian \(i\) is associated with mixture weights \(\mathbf{w}_i \in \mathbb{R}^L\) over the shared codebook \(\mathbf{E} \in \mathbb{R}^{L \times D}\). The decoded semantic feature \(\hat{\mathbf{f}}_i \in \mathbb{R}^D\) of Gaussian \(i\) and its \(L_2\)-normalized counterpart \(\tilde{\mathbf{f}}_i\) are given by:
\begin{equation}
\hat{\mathbf{f}}_i
=
\mathbf{E}^\top \mathbf{w}_i,
\qquad
\tilde{\mathbf{f}}_i
=
\frac{\hat{\mathbf{f}}_i}
{\left\|\hat{\mathbf{f}}_i\right\|_2 + \varepsilon}.
\end{equation}

To measure how well Gaussian \(i\) matches the prompt, we contrast the positive prompt embedding against each negative phrase:
\begin{equation}
r_i(p)
=
\min_{n \in \mathcal{N}}
\operatorname{softmax}\!\left(
\tau_c
\begin{bmatrix}
\tilde{\mathbf{f}}_i^{\top}\mathbf{t}^{+} \\
\tilde{\mathbf{f}}_i^{\top}\mathbf{t}_{n}^{-}
\end{bmatrix}
\right)_1,
\end{equation}
where \((\cdot)_1\) denotes the probability assigned to the positive prompt, and \(\tau_c = 10\) in our implementation. Intuitively, a Gaussian receives a high relevance score only if its decoded feature is more aligned with the prompt than with the generic negative phrases.

A prompt-specific Gaussian mask is then obtained by thresholding:
\begin{equation}
m_i(p)
=
\mathbb{I}\!\left[r_i(p) > \delta\right].
\end{equation}

For pixel-level prompt rendering, the renderer first splats the Gaussian mixture weights onto the image plane. Let \(\mathbf{w}(x) \in \mathbb{R}^L\) denote the rendered mixture weights at pixel \(x\). The semantic feature at that pixel is reconstructed and normalized as:
\begin{equation}
F(x)
=
\mathbf{E}^\top \mathbf{w}(x),
\qquad
\tilde{F}(x)
=
\frac{F(x)}
{\left\|F(x)\right\|_2 + \varepsilon}.
\end{equation}

Its prompt relevance is computed analogously:
\begin{equation}
R(x \mid p)
=
\min_{n \in \mathcal{N}}
\operatorname{softmax}\!\left(
\tau_c
\begin{bmatrix}
\tilde{F}(x)^{\top}\mathbf{t}^{+} \\
\tilde{F}(x)^{\top}\mathbf{t}_{n}^{-}
\end{bmatrix}
\right)_1.
\end{equation}

The resulting relevance map is used to highlight image regions that are most semantically aligned with the text prompt.

\section{Entropy-Adaptive Gaussian Sampling for Generative VLMs}
\label{sec:appendix_entropy_sampling}

As demonstrated by recent works such as SplatTalk~\cite{DBLP:journals/corr/abs-2503-06271}, projecting 3D Gaussian features into discrete tokens enables Large Language Models (LLMs) to perform zero-shot 3D visual question answering and scene understanding. However, directly passing all \(N_g\) Gaussians from an uncompressed SLAM map into the LLM backbone of a generative VLM is computationally prohibitive and yields highly redundant tokens.

Rather than relying on naive spatial down-sampling, we condense the scene into a compact sequence of \(M\) informative tokens (\(M \ll N_g\)) by ranking Gaussians according to the feature-channel entropy of their VQ-decoded semantic embeddings. This entropy is computed after VQ decoding and is not the entropy of the \textit{Perceiver}'s codeword-assignment weights.

Let \(\hat{\mathbf{f}}_i \in \mathbb{R}^D\) denote the decoded semantic embedding of Gaussian \(i\). For ranking, we apply a softmax across its \(D\) feature channels to obtain a normalized channel distribution \(q_{i,d}\):
\begin{equation}
q_{i,d} = \frac{\exp(\hat{f}_{i,d})}{\sum_{j=1}^D \exp(\hat{f}_{i,j})}.
\end{equation}

The feature-channel entropy \(H_i\) is then calculated as
\begin{equation}
H_i = - \sum_{d=1}^D q_{i,d} \log q_{i,d}.
\end{equation}

Low entropy indicates that the normalized response is concentrated in a small number of embedding channels, whereas high entropy indicates that it is distributed across more channels. We use high feature-channel entropy as a heuristic for selecting feature-rich Gaussian tokens.

To select the final \(M\) tokens, we rank the Gaussians by these entropy scores and extract the subset \(\mathcal{S}\) with the highest feature-channel entropy:
\begin{equation}
\mathcal{S} = \mathop{\operatorname{arg\,top-}\!M}_{i \in \{1, \dots, N_g\}} H_i.
\end{equation}

The resulting \(M\) Gaussians are then projected into the VLM's embedding space. This deterministic sampling reduces the token count while prioritizing decoded embeddings with high feature-channel entropy. The resulting token sequence bridges the continuous geometric scene representation with the VLM's discrete context window for tasks such as 3D visual question answering (VQA) and zero-shot scene captioning.

\section{Benchmark Details}
\label{sec:benchmark_details}

We provide benchmark details in Table~\ref{tab:dataset_summary}.

\section{Additional Qualitative Results}
\label{sec:appendix_qualitative}

We provide an additional qualitative example demonstrating the 3D scene captioning capabilities of X-GS in Figure~\ref{fig:captioning}.

\input{figures/a1_captioning_fig}

\section{Towards Embodied AI and World Models}
\label{sec:appendix_embodied}
Finally, as a future direction for embodied AI applications, the X-GS-\textit{Thinker} can be configured to interface with a Vision-Language-Action (VLA) model \cite{DBLP:journals/tnn/MaSZHK26, ma-etal-2025-astra, DBLP:journals/corr/abs-2501-16698}. By feeding language-aligned 3D geometric features directly into a VLA, our system provides the spatial information required to support embodied tasks.

Additionally, X-GS serves as a natural foundation for developing 3D-centric world models \cite{DBLP:journals/corr/abs-2603-19312}. Existing world models typically predict future states via computationally expensive 2D image or video generation \cite{DBLP:journals/corr/abs-2506-21539}. In contrast, a 3DGS-based world model operates with significantly greater efficiency; simulating an action such as object manipulation simply requires segmenting the relevant Gaussians and applying rigid affine transformations (e.g., translation and rotation)—an inherently lightweight computational operation \cite{DBLP:conf/cvpr/XieZQLF0J24, ma2026fourdgswam}.

%% file: tables/a1_benchmark.tex
\begin{table*}[t]
    \centering
    \caption{\textbf{Summary of benchmarking specifications.}}
    \label{tab:dataset_summary}
    \small
    \resizebox{1.0\textwidth}{!}{%
    \begin{tabular}{l c l c}
        \toprule
        \textbf{Dataset} & \textbf{\# Scenes} & \textbf{Scenes Evaluated} & \textbf{\# Queries} \\
        \midrule
        \multicolumn{4}{l}{\textit{\textbf{Online SLAM}}} \\
        TUM RGB-D~\cite{DBLP:conf/iros/SturmEEBC12} & 3 & \texttt{fr1\_desk, fr2\_xyz, fr3\_office} & \NA \\
        Replica~\cite{DBLP:journals/corr/abs-1906-05797} & 8 & \texttt{office0--4, room0--2} & \NA \\
        \midrule
        \multicolumn{4}{l}{\textit{\textbf{3D Visual Grounding}}} \\
        LERF~\cite{DBLP:conf/iccv/KerrKGKT23} & 4 & \texttt{teatime, figurines, ramen, waldo\_kitchen} & 67 (14--21/scene) \\
        3D-OVS~\cite{DBLP:conf/nips/LiuZZXYETXL23} & 5 & \texttt{bed, bench, lawn, room, sofa} & 31 ($\times$5 views) \\
        \midrule
        \multicolumn{4}{l}{\textit{\textbf{3D Vision-Language Reasoning}}} \\
        Scan2Cap~\cite{DBLP:conf/cvpr/ChenGNC21} (3D Captioning) & 800 & \texttt{ScanNet} (Validation set: 141) & 51K (9.5K eval) \\
        ScanQA~\cite{DBLP:conf/cvpr/AzumaMKK22} (3D QA) & 800 & \texttt{ScanNet} (Validation set: 71) & 41K (4.7K eval) \\
        SQA3D~\cite{DBLP:conf/iclr/MaYZ0LZH23} (Situated 3D QA) & 650 & \texttt{ScanNet} (Test set: 67) & 33K (3.5K eval) \\
        \bottomrule
    \end{tabular}
    }
\end{table*}

%% file: figures/a1_captioning_fig.tex
\begin{figure}[ht]
\centering
\includegraphics[width=1.0\columnwidth, trim={0.0in 10.1in 4.875in 0.25in},clip]{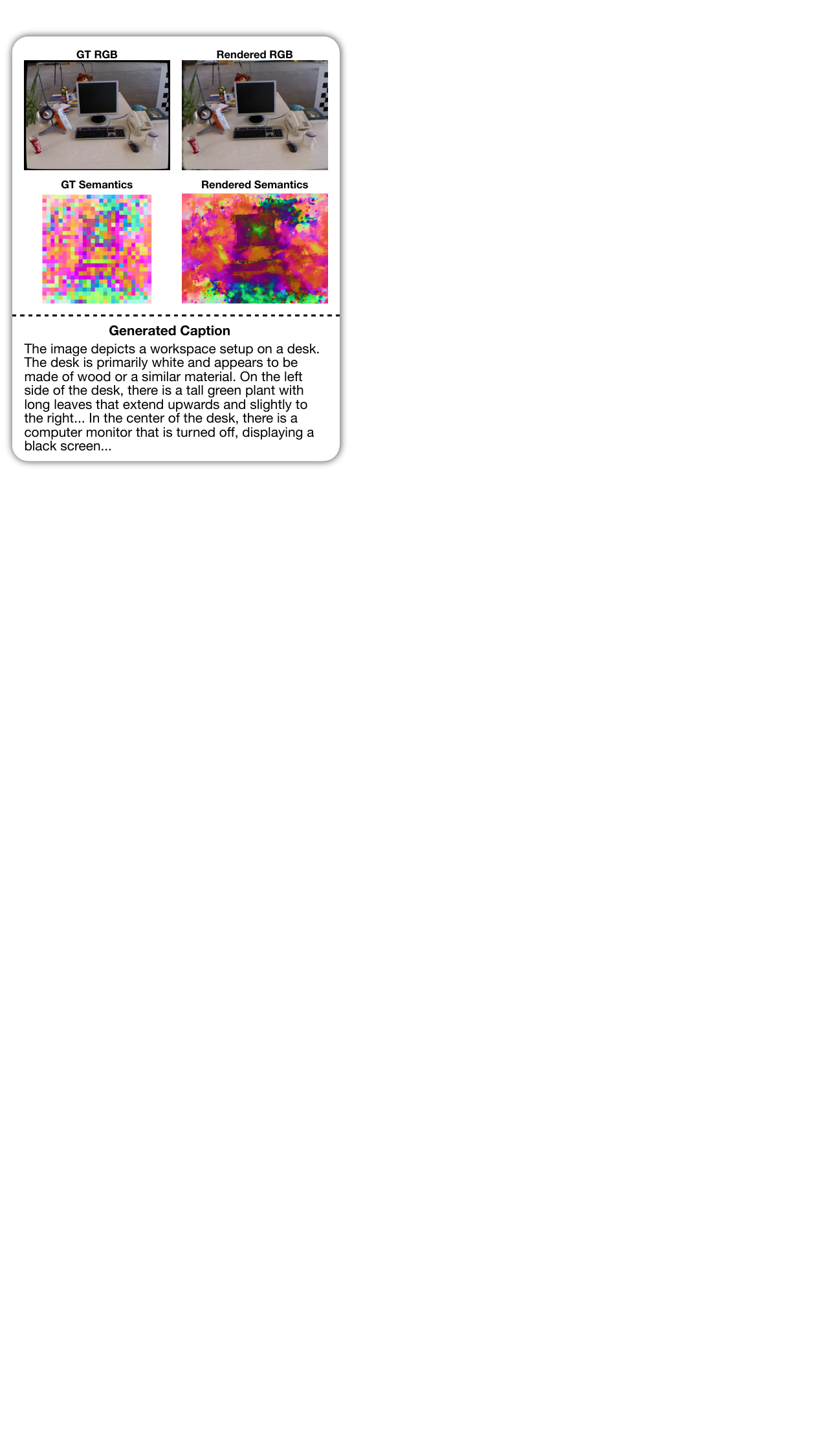}
\caption{Qualitative results of X-GS for 3D scene caption generation. The VLM generates captions in a zero-shot setting by taking 3D semantic Gaussians as input. Note that only a portion of the generated content is shown.}
\label{fig:captioning}
\end{figure}